%% file: main.tex
\documentclass{article}
\usepackage{spconf}
\usepackage{amsmath}
\usepackage{graphicx}
\usepackage{amssymb}
\usepackage{multirow}
\usepackage{array}
\usepackage{indentfirst}
\usepackage{svg}
\usepackage{float}
\usepackage[colorlinks,citecolor=green,urlcolor=blue,bookmarks=false,hypertexnames=true]{hyperref} 
\usepackage{pifont}
\usepackage{authblk}
\usepackage{mathabx}
\usepackage{algorithm}
\usepackage{algpseudocodex}
\usepackage{setspace}

\newcommand{\JandF}{\mathcal{J}\&\mathcal{F}}%
\newcommand{\cyan}[1]{\textcolor{cyan}{#1}}
\newcommand{\red}[1]{\textcolor{red}{#1}}
\newcommand{\cmark}{\text{\ding{51}}}%
\newcommand{\xmark}{\text{\ding{55}}}%

\title{TrickVOS: A Bag of Tricks for Video Object Segmentation}

\begin{document}
\name{Evangelos Skartados\textsuperscript{1}*,
Konstantinos Georgiadis\textsuperscript{1}*,
M. Kerim Yucel\textsuperscript{2}*\\ 
Koskinas Ioannis\textsuperscript{1}, Armando Domi\textsuperscript{1}, Anastasios Drosou\textsuperscript{1}, Bruno Manganelli\textsuperscript{2},
Albert Sa\`a-Garriga\textsuperscript{2}\thanks{* The first three authors contributed equally.}}
\address{\textsuperscript{1} CERTH, Information Technologies Institute, Thessaloniki, Greece \\
\textsuperscript{2} Samsung Research UK}
\maketitle
\begin{abstract}
 Space-time memory (STM) network methods have been  dominant in semi-supervised video object segmentation (SVOS) due to their remarkable performance. In this work, we identify three key aspects where we can improve such methods; i) supervisory signal, ii) pretraining and iii) spatial awareness. We then propose \textbf{TrickVOS}; a generic, method-agnostic bag of tricks addressing each aspect with i) a structure-aware hybrid loss, ii) a simple decoder pretraining regime and iii) a cheap tracker that imposes spatial constraints in model predictions. Finally, we propose a lightweight network and show that when trained with TrickVOS, it achieves competitive results to state-of-the-art methods on DAVIS and YouTube benchmarks, while being one of the first STM-based SVOS methods that can run in real-time on a mobile device.
 
\end{abstract}

\vspace{-3mm}

\input{introduction}

\input{related_work}
\input{methodology}

\input{experiments}

\input{conclusion}

\clearpage

\begingroup
\setstretch{0.95}
\bibliographystyle{IEEEbib}
\bibliography{strings,refs}
\endgroup

\end{document}

%% file: introduction.tex
\section{Introduction} \label{introduction}
\vspace{-4mm}
\noindent  Semi-supervised video object segmentation segments objects, where object mask annotations are given only for the first frame. SVOS is the backbone of many applications, such as surveillance, autonomous driving and video manipulation. It is a difficult task, where robustness against occlusions, shape changes, motion and distractors are necessary, and performing well irrespective of the object classes is imperative. In recent years, space-time memory networks have gained traction \cite{oh2019video,cheng2021rethinking,li2022recurrent,wang2021swiftnet}, as they have made significant advances. Nevertheless, there is still a considerable room for improvement.  

In this paper, we aim to improve STM networks by introducing a generic, method-agnostic bag-of-tricks. We first identify three points relatively unexplored in the literature; i) supervisory signals, ii) pretraining and iii) spatial constraints. We then propose TrickVOS that includes solutions for each of these points. The contributions of TrickVOS are as follows; i) a training loss that leverages the structural information available in the ground-truth masks, with an emphasis on the tracked object regions, ii) a simple pretraining regime that addresses the problem of not being able to effectively pretrain decoders and iii) a cheap tracker that filters the model predictions based on historical motion patterns. Each trick improves the results, and their combination leads to further improvements. Finally, we demonstrate that TrickVOS makes a lightweight network competitive to significantly larger/expensive methods on multiple SVOS benchmarks. TrickVOS is visualized in Figure \ref{fig:overall_diagram}.

\vspace{-1mm}

%% file: related_work.tex
\vspace{-3mm}
\section{Related Work} \label{related_work}
\vspace{-3mm}

\noindent \textbf{Semi-supervised Video Object Segmentation.} SVOS methods span a large spectrum, where per-mask online finetuning \cite{li2018video}, flow-based \cite{tsai2016video}, detection-aided \cite{luiten2019premvos} and matching-based methods are prominent examples. STM based methods have been the dominant approach, where a feature correspondence problem is solved between the current frame and the frames stored in the memory bank to predict the object masks \cite{oh2019video,cheng2021rethinking,li2022recurrent,wang2021swiftnet}. Follow-up works have focused on increasing efficiency \cite{wang2021swiftnet,cheng2021rethinking}, accuracy on long-videos \cite{cheng2022xmem}, bounding the memory limit \cite{li2022recurrent,miles2023mobilevos} and eliminating spatial redundancy \cite{miao2022region,xie2021efficient}. Despite these advances, there are unexplored areas where STM based SVOS methods can be further improved.

The tricks of TrickVOS can be easily plugged into other methods. To the best of our knowledge, we are the first to propose an enhanced loss function and a decoder pretraining regime to improve the accuracy of SVOS methods. Similar ideas for spatial-filtering have been explored in \cite{xie2021efficient,miao2022region}, but ours have key differences; we i) do not use a flow network as \cite{xie2021efficient} or ii) do not perform filtering on the feature maps as \cite{xie2021efficient,miao2022region}. TrickVOS paves the way for accurate STM-based SVOS models that can run in real-time on mobile devices.

\vspace{-4mm}

%% file: methodology.tex
\section{TrickVOS}
\label{methodology}
\vspace{-4mm}
\subsection{SVOS Preliminaries}
\vspace{-2mm}

\begin{figure*}[!ht]
  \centering
        \includegraphics[width=0.95\textwidth]{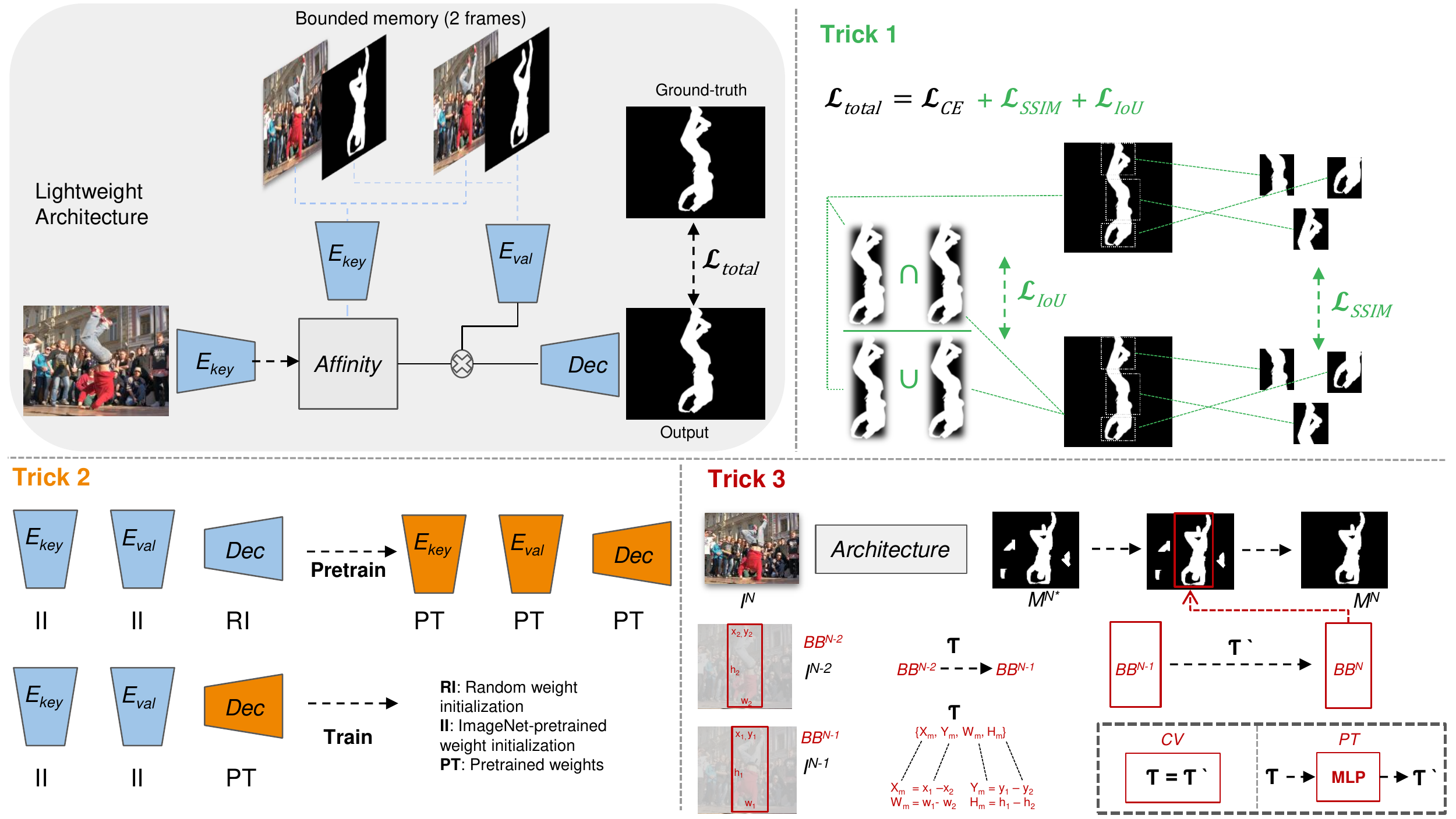}
  \vspace{-4mm}
  \caption{TrickVOS is formed of three tricks that improve semi-supervised video object segmentation; Trick 1 introduces a loss that promotes foreground focus and structure-awareness, Trick 2 presents a simple decoder pretraining scheme and Trick 3 filters predictions via a cheap tracker with two variants, constant velocity (CV) and parametric tracker (PT). } 
  \label{fig:overall_diagram}
  \vspace{-6mm}
\end{figure*}

\noindent Given a set of $K$ object masks $M_{obj}\in\mathbb{N}^{HW}$, $M=\{M_{1}, ..., M_{k}\}$ in the first frame $I^{1}\in\mathbb{R}^{HW\times 3}$ of an N-frame long sequence $I= \{I^{1}, .., I^{N}  \}$, SVOS aims to track these objects in the sequence. We now identify key issues and propose solutions to improve the performance of SVOS methods.

\vspace{-3mm}
\subsection{Trick 1: A Better Loss Function}
\vspace{-2mm}
\noindent \textbf{Problem.} SVOS is a dense classification task, where each pixel is classified as background or an object. Therefore, most methods train with a form of cross-entropy (CE) loss \cite{cheng2021rethinking}. However, SVOS is highly sensitive to object shapes, which requires structural information. Structural information also provides boundary awareness, where most SVOS prediction errors tend to occur. Additionally, during training, foreground focus is highly desirable for accuracy and rapid convergence. Neither structural information, nor foreground-focus are provided by CE since it considers all pixels as independent.    

\noindent \textbf{Our solution.} Our first trick aims to provide structural information and foreground-focus during training. For the former, a natural choice is the SSIM metric \cite{wang2003multiscale}. SSIM can be used as a loss function to promote structural similarity between predictions and ground-truths, and since its patch-level nature provides rich context information, it might help the network to model the appearance/shape changes more effectively. For the latter, we hypothesize that the Jaccard-index, namely Intersection-over-Union (IoU), can be helpful as it measures the alignment between the predicted and the ground-truth masks. The alignment of the masks naturally focus on the foreground, therefore it meets our criteria. Various differentiable IoU formulations are used in many tasks \cite{mattyus2017deeproadmapper}, including SVOS \cite{yu2022batman}, as loss functions. Using the IoU loss of \cite{mattyus2017deeproadmapper}, we define the total loss as \cite{qin2019basnet} 
\vspace{-1mm}
\begin{equation}
    \mathcal{L}_{total} = \lambda_{1} \mathcal{L}_{CE} + 
 \lambda_{2} \mathcal{L}_{SSIM} + \lambda_{3} \mathcal{L}_{IoU} 
 \vspace{-1.5mm}
\end{equation}

\noindent where $\lambda_{1}=\lambda_{2}=\lambda_{3}=1$. This formulation is shown to be effective in salient object detection \cite{qin2019basnet}, which is somewhat similar to SVOS. See Figure \ref{fig:overall_diagram} for a diagram of Trick 1.
\vspace{-5mm}

\subsection{Trick 2: Decoder Pretraining}
\vspace{-2mm}
\noindent \textbf{Problem.} Using ImageNet-pretrained encoders in dense prediction tasks is a ubiquitous practice, as it often improves the downstream accuracy. A similar pretraining for decoders is not straightforward; decoders directly produce the final result, which differs for various tasks (i.e. depth vs saliency). Therefore, a decoder trained on a task might not be a good initialization for another task. Self-supervised learning \cite{brempong2022denoising} and generative pretraining \cite{zoph2020rethinking} are shown to beat random initialization for decoders, but none of them target SVOS. 

\noindent \textbf{Our solution.} Our second trick is probably the simplest one; we simply re-use a decoder that is already trained on our task. We train a model through all the stages and save the decoder weights. Later, we train a model with random weight initialization, except we initialize the decoder weights from the previous training. "Why not train the whole model for longer instead?", one might ask. Alternatively, one might question as to why we do not do the pretraining for the whole model,  not just the decoder. We observe that in both cases, the model overfits to DAVIS and performs poorly on YouTube. Our decoder pretraining, however, consistently improves on all datasets (see Table \ref{tab:ablation}). We test alternative strategies \cite{brempong2022denoising} but observe poor results. See Figure \ref{fig:overall_diagram} for a diagram of Trick 2. 

\vspace{-4mm}
\setlength{\textfloatsep}{1pt}
\begin{algorithm}
 \scriptsize
  \caption{Trick 3 pseudocode.}
  \begin{algorithmic}[1]
    \Function{Trick3}{$I,M$} \Comment{Inputs: Frames I and predicted masks M.} 
      \State C  = 1
      \While{$C\not=len(I)$}
        \If{$C<3$}
            \State $BB^{C} = init(M^{C})$ \Comment{Initialize bbox from mask.} 
            \State $C+=1$
            \State \textbf{continue}
        \Else{}
            \State $\mathcal{T}=BB^{C-1} - BB^{C-2}$ \Comment{Historical motion pattern.}
                \If{$CV$}
                    \State $\mathcal{T'}=\mathcal{T}$ \Comment{Constant Velocity variant.}
                \Else{}
                    \State $\mathcal{T'}=MLP(\mathcal{T})$
                    \Comment{Parametric Tracker variant.}
                \EndIf
        \EndIf
        \State $BB^{C} = \mathcal{T'} +  BB^{C-1}$ \Comment{Apply motion to predict the bounding box.}
        \State $AM^{C} \gets BB^{C}$ \Comment{Attention map from bounding box.}
        \State $M'^{C} = AM^{C} \times M^{C}$ \Comment{Filtered mask.}
        \State $BB^{C} = init(M'^{C})$ \Comment{Reinitialize bbox from mask.}
        \State $C+=1$
      \EndWhile
      \State $M' = \{M'^{1}, .., M'^{C}  \}$
      \State \textbf{return} $M'$ \Comment{Output: Filtered masks.}
      \EndFunction
  \end{algorithmic} \label{algo:trick_3}
\end{algorithm}

\subsection{Trick 3: Prediction Filtering}
\vspace{-2mm}
\noindent \textbf{Problem.} STM-based methods perform global feature matching between current and memory frames, which might lead to erroneous matches, especially when objects of same types/appearance are present in the frame. Some alternatives tackle this issue up to a degree, but they require relevant architectural changes since they predict bounding boxes on features \cite{miao2022region} or use expensive optical flow \cite{xie2021efficient}. 

\noindent \textbf{Our solution.} Our final trick is a tracker that filters the predictions. The core idea is simple; we track a bounding box throughout the video that encapsulates an object mask and filter out predictions outside the area of the bounding box. We propose two variants of our tracker, Constant Velocity (CV) and Parametric Tracker (PT). Trick 3 is shown in Figure \ref{fig:overall_diagram}.

\noindent \textbf{Variant 1.} CV is extremely simple; it is not a tracker at all. We simply assume constant velocity for objects; the motion $\mathcal{T}$  between the previous two bounding boxes is used to calculate the bounding box of the next frame. This bounding box is converted to an attention map, multiplied by the predicted mask to generate the filtered mask. Note that $\mathcal{T}$ is defined as the change in top left coordinates, width and height.

\noindent \textbf{Variant 2.} PT extends CV by using an MLP to predict the next bounding box. Instead of assuming a constant $\mathcal{T}$, PT feeds the motion between two previous frames into a 5-layer MLP (with GeLU activations in between) to predict the next bounding box.  Both CV and PT offer a great deal of flexibility, as CV/PT can be plugged into any method or architecture. See Algorithm \ref{algo:trick_3} for a detailed breakdown of Trick 3.

\vspace{-5mm}

%% file: experiments.tex
\section{Experiments} \label{sec:experiments}
\vspace{-3mm}
\subsection{Experimental Details}
\vspace{-2mm}
\noindent \textbf{Architecture.} We adopt the STCN architecture \cite{cheng2021rethinking} with key changes; we i) fix the memory size to two frames (first and last) to avoid memory explosion, ii) replace key/value encoders with MobileNetV2 \cite{sandler2018mobilenetv2}, iii) reduce key dimension to 32, iv) remove ASPP and v) use mask as the only input to the value encoder \cite{miles2023mobilevos}. These changes make our architecture, denoted as the \textit{baseline} throughout the paper, less accurate than STCN, but improves its runtime performance (see Figure \ref{fig:overall_diagram}).

\noindent \textbf{Datasets.} DAVIS 2016 \cite{perazzi2016benchmark} is a single-object SVOS benchmark with 30 training and 20 validation videos, where DAVIS 2017 \cite{pont20172017} extends it to multi-object with 60 training videos, 30 test and 30 validation videos. YouTube-VOS 2019 \cite{xu2018youtube} is a large-scale multi-object SVOS benchmark with 3471 training videos formed of 65 categories and 507 videos for validation. The validation set includes 26 unseen object categories, allowing model assessment on unseen object categories. 

\noindent \textbf{Metrics.} We use the metrics region similarity $\mathcal{J}$ and contour accuracy $\mathcal{F}$, and their average $\mathcal{J} \& \mathcal{F}$. On YouTube, we also report seen and unseen class $\mathcal{J} \& \mathcal{F}$. We evaluate our methods on the validation sets of DAVIS and YouTube datasets.

\noindent \textbf{Training Details.}
We follow the training regime of \cite{cheng2021rethinking}, where we first pretrain on static images with synthetic deformations, and then train on YouTube and DAVIS 2017 jointly for 300K iterations with a batch size of 16. We use Adam optimizer with a learning rate of $1e-5$ and a weight decay of $1e-7$. The only difference is that we use a sequence length of 5 instead of 3. Trick 3 PT variant is trained separately, using bounding boxes extracted from the ground-truth masks. In PT training, we further penalize the predicted bounding boxes that are smaller than the ground-truth by increasing their loss weights. We use RTX 3090 GPUs for all experiments.

\vspace{-4mm}
\subsection{Ablation Studies}
\vspace{-2mm}
\noindent We conduct detailed ablation studies to show the effect of each TrickVOS component.
The results are shown in Table \ref{tab:ablation}. We use a single model checkpoint to evaluate on all datasets.

 \begin{table}[t!]
\resizebox{0.48\textwidth}{!}{%
\begin{tabular}{l|c|c|c}
Method & DAVIS'16 & DAVIS'17 & YouTube-VOS'19 \\ \hline
Baseline & 88.18 & 80.86 & 77.18 	\\
+  SSIM  & 87.81 \red{$\downarrow$0.37} & 81.25 \cyan{$\uparrow$0.39} & 78.42 \cyan{$\uparrow$1.24}	\\
+  IoU  & 89.04 \cyan{$\uparrow$0.86} & 81.88 \cyan{$\uparrow$1.02} & 79.39 \cyan{$\uparrow$2.21}\\
+  Trick 1 & 88.80 \cyan{$\uparrow$0.62}& 81.97 \cyan{$\uparrow$1.11}  & 79.42 \cyan{$\uparrow$2.24}	\\ \hline
+ Trick 2 & 88.47 \cyan{$\uparrow$0.25} & 81.06 \cyan{$\uparrow$0.20} & 	78.07 \cyan{$\uparrow$0.89}\\ \hline
+ Trick 3 (CV) & 88.30 \cyan{$\uparrow$0.12} & 81.17 \cyan{$\uparrow$0.31} & 78.04 \cyan{$\uparrow$0.86}	\\ 
+ Trick 3 (PT) & 88.31 \cyan{$\uparrow$0.13} & 81.23 \cyan{$\uparrow$0.37} & 78.15 \cyan{$\uparrow$0.97}	\\ \hline
+ Trick 1+2 & 89.19 \cyan{$\uparrow$0.99} &  82.54 \cyan{$\uparrow$1.68} & 79.94 \cyan{$\uparrow$2.76}	\\
+ TrickVOS (CV) & 89.28 \cyan{$\uparrow$1.10} &  82.56 \cyan{$\uparrow$1.70} & 80.38 \cyan{$\uparrow$3.20}	\\
+ TrickVOS (PT) & 89.29 \cyan{$\uparrow$1.11} & 82.67 \cyan{$\uparrow$1.81} & 80.47 \cyan{$\uparrow$3.29}	\\

\end{tabular}%
}
\vspace{-4mm}
    \caption{Ablation study $\mathcal{J} \& \mathcal{F}$ values. CV and PT indicate constant velocity and parametric variants of our tracker (Trick 3). TrickVOS combines all tricks. \cyan{$\uparrow$} and \red{$\downarrow$} show $\mathcal{J} \& \mathcal{F}$ increase and decrease compared to the baseline, respectively. }
\vspace{1mm}
    \label{tab:ablation}
\end{table}

\begin{table*}[t!]
\resizebox{\textwidth}{!}{%
\begin{tabular}{lc|ccccccccccccc}
\multicolumn{2}{l|}{Method} & \multicolumn{3}{c|}{DAVIS'16} & \multicolumn{3}{c|}{DAVIS'17} & \multicolumn{5}{c|}{YouTube'19} & \multicolumn{2}{c}{FPS} \\ \hline
\multicolumn{1}{c}{} & \multicolumn{1}{c|}{CC} & $\mathcal{J} \& \mathcal{F}$ & $\mathcal{J}$ & \multicolumn{1}{c|}{$\mathcal{F}$} & $\mathcal{J} \& \mathcal{F}$& $\mathcal{J}$ & \multicolumn{1}{c|}{$\mathcal{F}$} & $\mathcal{J} \& \mathcal{F}$ & $\mathcal{J}_{seen}$ & $\mathcal{F}_{seen}$ & $\mathcal{J}_{unseen}$ & \multicolumn{1}{c|}{$\mathcal{F}_{unseen}$} & DAVIS'16 & DAVIS'17 \\ \hline
MiVOS$\ddagger $\cite{cheng2021modular} & $\xmark$ & 91.0 & 89.7 & \multicolumn{1}{c|}{92.4} & 84.5 & 81.7 & \multicolumn{1}{c|}{87.4} & 82.4 & 80.6 & 84.7 & 78.2 & \multicolumn{1}{c|}{85.9} & 16.9 & 11.2 \\
STM \cite{oh2019video} & $\xmark$ & 89.3 & 88.7 & \multicolumn{1}{c|}{89.9} & 81.8 & 79.2 & \multicolumn{1}{c|}{84.3} & 79.2 & 79.6 & 83.6 & 73.0 & \multicolumn{1}{c|}{80.6} & 6.30 & 10.2 \\
STCN$\ddagger$ \cite{cheng2021rethinking} & $\xmark$ & 91.7 & 90.4 & \multicolumn{1}{c|}{93.0} & 85.3 & 82.0 & \multicolumn{1}{c|}{88.6} & 84.2 & 82.6 & 87.0 & 79.4 & \multicolumn{1}{c|}{87.7} & 26.9 & 20.2 \\
XMem \cite{cheng2022xmem} & $\xmark$  & 91.5 & 90.4 & \multicolumn{1}{c|}{92.7} & 86.2 & 82.9 & \multicolumn{1}{c|}{89.5} & 85.5 & 84.3 & 88.6 & 80.3 & \multicolumn{1}{c|}{88.6} & 29.6 & 22.6 \\ \hline
GCNet \cite{li2020fast} & $\cmark$ & 86.6 & 87.6 & \multicolumn{1}{c|}{85.7} & 71.4 & 69.3 & \multicolumn{1}{c|}{73.5} & 73.2 & 72.6 & 75.6 & 68.9 & \multicolumn{1}{c|}{75.7} & 25.0 & \textless{}25.0 \\
SwiftNet \cite{wang2021swiftnet} & $\cmark$ & 90.4 & 90.5 & \multicolumn{1}{c|}{90.3} & 81.1 & 78.3 & \multicolumn{1}{c|}{83.9} & 77.8 & 77.8 & 81.8 & 72.3 & \multicolumn{1}{c|}{79.5} & 25.0 & \textless{}25.0 \\
CFBI  \cite{yang2020collaborative} & $\cmark$ & 89.9 & 88.7 & \multicolumn{1}{c|}{91.1} & 81.9 & 79.3 & \multicolumn{1}{c|}{84.5} & 81.0 & 80.6 & 85.1 & 75.2 & \multicolumn{1}{c|}{83.0} & 5.90 & -- \\
RDE-VOS \cite{li2022recurrent} & $\cmark$ & 91.1 & 89.7 & \multicolumn{1}{c|}{92.5} & 84.2 & 80.8 & \multicolumn{1}{c|}{87.5} & 81.9 & 81.1 & 85.5 & 76.2 & \multicolumn{1}{c|}{84.8} & 35.0 & 27.0 \\ \hline
Baseline & $\cmark$ & 88.2 & 87.5 & \multicolumn{1}{c|}{88.8} & 80.9 & 77.5 & \multicolumn{1}{c|}{84.2} & 77.2 & 76.9 & 81.0 & 71.3 & \multicolumn{1}{c|}{79.4} & {102.1} & {99.5} \\
Baseline + \textbf{TrickVOS (CV)} & $\cmark$ & 89.3 & 88.7 & \multicolumn{1}{c|}{89.8} & 82.6 & 79.3 & \multicolumn{1}{c|}{85.9} & 80.3 & 80.0 & 83.9 & 74.5 & \multicolumn{1}{c|}{82.9} & 91.4  & 79.3 \\
Baseline + \textbf{TrickVOS (PT)} & $\cmark$ & 89.3 & 88.7 & \multicolumn{1}{c|}{89.9} & 82.7 &  79.4 & \multicolumn{1}{c|}{86.0} & 80.5 & 79.5 & 83.3 & 75.2 & \multicolumn{1}{c|}{84.0} & 86.4 & 76.4 \\ \hline
STCN $\dagger$ \cite{cheng2021rethinking} & $\xmark$ & 91.2 & 90.1 & \multicolumn{1}{c|}{92.3} & 84.0 & 80.6 & \multicolumn{1}{c|}{87.4} & 80.7 & 80.2 & 84.5 & 75.6 & \multicolumn{1}{c|}{82.7} & 62.4 & 48.3 \\
STCN $\dagger$ \cite{cheng2021rethinking} + \textbf{Trick 1+2} & $\xmark$ & 91.7 & 90.4 & \multicolumn{1}{c|}{92.9} & 85.4 & 82.1 & \multicolumn{1}{c|}{88.8} & 81.6 & 81.2 & 85.5 & 75.8 & \multicolumn{1}{c|}{83.9} & 62.4 & 48.3 \\
STCN $\dagger$ \cite{cheng2021rethinking} + \textbf{TrickVOS (PT)} & $\xmark$ & 91.8 & 90.5 & \multicolumn{1}{c|}{93.1} & 86.1 & 82.6 & \multicolumn{1}{c|}{89.6} & 82.8 & 82.1 & 86.4 & 77.2 & \multicolumn{1}{c|}{85.5} & 45.4 & 35.1 \\
\end{tabular}%
}
\vspace{-4mm}
    \caption{Results on the DAVIS 2016/2017 and YouTube-VOS 2019 validation sets. \textbf{CC} indicate constant cost. 
    $\ddagger$ indicates models trained with additional data (BL30K). $\dagger$ indicates STCN re-training by us (without ASPP module or BL30K pretraining).
    }
    \label{tab:sota_results}
    \vspace{-6mm}
\end{table*}

\noindent \textbf{Trick 1.} We add SSIM and IoU losses over the cross-entropy baseline, and then combine all of them. Table \ref{tab:ablation} third row shows that IoU increases the accuracy significantly. The second row shows that SSIM improves DAVIS'17 but hurts DAVIS'16. Note that we focus on DAVIS'17 as it is a larger dataset with multiple object annotations, which is more reminiscent of an actual SVOS use case. When we combine all (Table \ref{tab:ablation} 4th row), we see solid improvements; 0.62, 1.11  and 2.24 $\mathcal{J} \& \mathcal{F}$ on DAVIS'16, 17 and YouTube, respectively.

\noindent \textbf{Trick 2.} Fifth row of Table \ref{tab:ablation} (Trick 2) shows that we get consistent improvements on all datasets. Note that this result shows that we do not overfit, especially to DAVIS since it is quite smaller compared to YouTube. Improvements on YouTube especially emphasize the benefit of Trick 2.

\noindent \textbf{Trick 3.} We observe improvements even with the constant velocity variant (Table \ref{tab:ablation} row 6, Trick 3 CV), which has little computational overhead. With the PT variant, we observe slightly better results compared to CV, but at the cost of a slightly increased performance overhead. Note that we do not explicitly optimize the tracker, therefore the performance overhead can be reduced with optimized implementations.

\noindent \textbf{All tricks.} An important question remains; can we simply use all tricks at the same time and get further improvement? The results show that Trick 1 and 2 combined perform better on all datasets, compared to using them separately. When all tricks are combined, we get the largest improvement on all datasets.

\vspace{-5mm}
    
\subsection{Results}
\vspace{-2mm}
\noindent We compare TrickVOS against the state-of-the-art methods. We indicate methods with bounded memory during inference with Constant Cost (CC) \cite{li2022recurrent}. Unlike others \cite{li2022recurrent}, we use a single model checkpoint for all dataset evaluations. The results are shown in Table \ref{tab:sota_results}. Note that methods with unbounded memory are not practical, as they eventually run out of memory and their FPS decrease. We provide their results, but we mainly compare ours against other constant cost methods.

\noindent \textbf{DAVIS 2016.} Our model performs quite competitively; it is slightly behind RDE-VOS, but it is $\times$2.5 faster and has $\times$32 fewer parameters (our 1.9M vs their 64M). Our methods comfortably perform in real-time, with minimal reductions (10\% and 15\% for CV and PT) in runtime compared to the baseline.

\noindent \textbf{DAVIS 2017.} In multi-object scenarios, the results look even better. We are the second best after RDE-VOS in constant cost methods. Also note that in multi-object scenarios, most of the methods perform significantly slower, but we are affected less than the others; we are now $\times$3 faster than RDE-VOS. Our methods are still comfortably in real-time regime, with acceptable reductions in runtime compared to the baseline.

\noindent \textbf{YouTube-VOS.} We observe the most significant improvements over our baseline on YouTube (3.3 $\JandF$), which emphasizes the benefit of our tricks as YouTube is the largest and the most diverse of all datasets we experiment with. 

\noindent \textbf{On-device deployment.} TrickVOS unlocks a practical application; mobile deployment. To test on-device performance, we convert our models to tflite and benchmark them using the tflite benchmarking tool. Our method reaches 30+ FPS on a Samsung Galaxy S22 GPU; to the best of our knowledge, this is the first STM-based SVOS method that achieves real-time on a mobile device. Note that we are already faster than others on high-end GPUs (see Table \ref{tab:sota_results}); assuming a naive linear scaling of speed from desktop to mobile GPUs and based on our 30 FPS, no other method is likely to perform in real-time.

\begin{figure}[!t]
  \centering
        \includegraphics[width=0.48\textwidth]{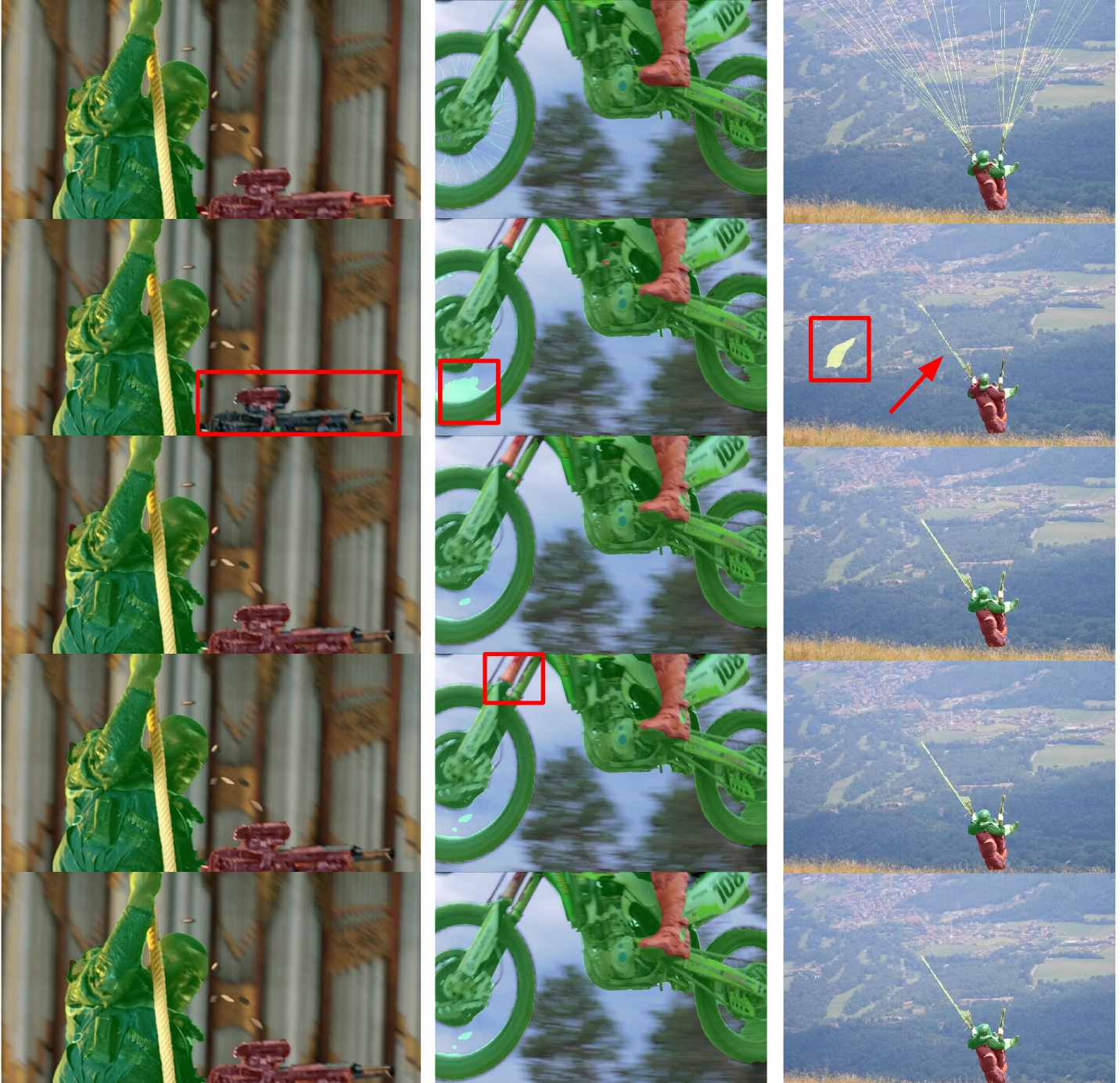}
  \vspace{-7mm}
  \caption{Rows show (top$\rightarrow$bottom) the ground-truth, baseline, baseline + trick 1, baseline + trick 2 and baseline + TrickVOS predictions. Note the red boxes/arrows showing the erroneous regions, which are gradually improved with our tricks. } 
  \vspace{1mm}
  \label{fig:qual}
\end{figure}

\noindent \textbf{Qualitative results.} Figure \ref{fig:qual} shows example images from DAVIS'17 validation set; the baseline fails to either track objects (gun in leftmost image) or mispredicts masks (erroneous predictions in the wheel and on the mountain). Our tricks, when applied, visibly improve such error cases.

\noindent \textbf{TrickVOS + other methods.} We apply TrickVOS to the original STCN (with no ASPP module or BL30K pretraining) to highlight its plug-and-play nature. For a fair comparison, we re-train it (using authors' code) with and without TrickVOS; the last three rows of Table \ref{tab:sota_results} show TrickVOS consistently improves the results on all datasets.

\vspace{-5mm}

%% file: conclusion.tex
\section{Conclusion}
\label{conclusion}
\vspace{-3mm}
We present TrickVOS, a model-agnostic bag of tricks that improves the accuracy of semi-supervised video object segmentation. Each trick introduces improvements, and when combined, they perform even better. TrickVOS, when applied to a lightweight architecture, performs competitively to the state-of-the-art, while running real-time on a mobile phone.

%% file: main.bbl
\begin{thebibliography}{10}

\bibitem{oh2019video}
Seoung~Wug Oh, Joon-Young Lee, Ning Xu, and Seon~Joo Kim,
\newblock ``Video object segmentation using space-time memory networks,''
\newblock in {\em International Conference on Computer Vision (ICCV)}, 2019,
  pp. 9226--9235.

\bibitem{cheng2021rethinking}
Ho~Kei Cheng, Yu-Wing Tai, and Chi-Keung Tang,
\newblock ``Rethinking space-time networks with improved memory coverage for
  efficient video object segmentation,''
\newblock {\em NeurIPS}, vol. 34, pp. 11781--11794, 2021.

\bibitem{li2022recurrent}
M.~Li, Li~Hu, Z.~Xiong, Bang Zhang, Pan Pan, and Dong Liu,
\newblock ``Recurrent dynamic embedding for video object segmentation,''
\newblock in {\em Conference on Computer Vision and Pattern Recognition
  (CVPR)}, 2022, pp. 1332--1341.

\bibitem{wang2021swiftnet}
Haochen Wang, Xiaolong Jiang, Haibing Ren, Yao Hu, and Song Bai,
\newblock ``Swiftnet: Real-time video object segmentation,''
\newblock in {\em Conference on Computer Vision and Pattern Recognition
  (CVPR)}, 2021, pp. 1296--1305.

\bibitem{li2018video}
Xiaoxiao Li and Chen~Change Loy,
\newblock ``Video object segmentation with joint re-identification and
  attention-aware mask propagation,''
\newblock in {\em European conference on computer vision (ECCV)}, 2018, pp.
  90--105.

\bibitem{tsai2016video}
YH~Tsai, MH~Yang, and MJ~Black,
\newblock ``Video segmentation via object flow,''
\newblock in {\em Conference on Computer Vision and Pattern Recognition
  (CVPR)}, 2016, pp. 3899--3908.

\bibitem{luiten2019premvos}
Jonathon Luiten, Paul Voigtlaender, and Bastian Leibe,
\newblock ``Premvos: Proposal-generation, refinement and merging for video
  object segmentation,''
\newblock in {\em Asian Conference on Computer Vision (ACCV)}, 2019, pp.
  565--580.

\bibitem{cheng2022xmem}
Ho~Kei Cheng and Alexander~G Schwing,
\newblock ``Xmem: Long-term video object segmentation with an atkinson-shiffrin
  memory model,''
\newblock in {\em European conference on computer vision (ECCV)}, 2022, pp.
  640--658.

\bibitem{miles2023mobilevos}
Roy Miles, Mehmet~Kerim Yucel, Bruno Manganelli, and Albert Sa{\`a}-Garriga,
\newblock ``Mobilevos: Real-time video object segmentation contrastive learning
  meets knowledge distillation,''
\newblock in {\em Proceedings of the IEEE/CVF Conference on Computer Vision and
  Pattern Recognition}, 2023, pp. 10480--10490.

\bibitem{miao2022region}
Bo~Miao, M.~Bennamoun, Yongsheng Gao, and Ajmal Mian,
\newblock ``Region aware video object segmentation with deep motion modeling,''
\newblock {\em arXiv:2207.10258}, 2022.

\bibitem{xie2021efficient}
H.~Xie, H.~Yao, S.~Zhou, S.~Zhang, and W.~Sun,
\newblock ``Efficient regional memory network for video object segmentation,''
\newblock in {\em Conference on Computer Vision and Pattern Recognition
  (CVPR)}, 2021, pp. 1286--1295.

\bibitem{wang2003multiscale}
Zhou Wang, Eero~P Simoncelli, and Alan~C Bovik,
\newblock ``Multiscale structural similarity for image quality assessment,''
\newblock in {\em ACSSC}. Ieee, 2003, vol.~2, pp. 1398--1402.

\bibitem{mattyus2017deeproadmapper}
Gell{\'e}rt M{\'a}ttyus, Wenjie Luo, and Raquel Urtasun,
\newblock ``Deeproadmapper: Extracting road topology from aerial images,''
\newblock in {\em International Conference on Computer Vision (ICCV)}, 2017,
  pp. 3438--3446.

\bibitem{yu2022batman}
Ye~Yu, J.~Yuan, G.~Mittal, Li~Fuxin, and Mei Chen,
\newblock ``Batman: Bilateral attention transformer in motion-appearance
  neighboring space for video object segmentation,''
\newblock in {\em European conference on computer vision (ECCV)}, 2022, pp.
  612--629.

\bibitem{qin2019basnet}
Xuebin Qin, Zichen Zhang, Chenyang Huang, Chao Gao, Masood Dehghan, and Martin
  Jagersand,
\newblock ``Basnet: Boundary-aware salient object detection,''
\newblock in {\em Conference on Computer Vision and Pattern Recognition
  (CVPR)}, 2019, pp. 7479--7489.

\bibitem{brempong2022denoising}
E.~A. Brempong, Simon K., Ting C., Niki P., Matthias M., and M.~N.,
\newblock ``Denoising pretraining for semantic segmentation,''
\newblock in {\em Conference on Computer Vision and Pattern Recognition
  (CVPR)}, 2022, pp. 4175--4186.

\bibitem{zoph2020rethinking}
Barret Z., Golnaz G., Tsung-Yi Lin, Yin Cui, H.~Liu, Ekin DC, and Quoc Le,
\newblock ``Rethinking pre-training and self-training,''
\newblock {\em NeurIPS}, vol. 33, pp. 3833--3845, 2020.

\bibitem{sandler2018mobilenetv2}
Mark S., Andrew H., M.~Zhu, Andrey Z., and LC~Chen,
\newblock ``Mobilenetv2: Inverted residuals and linear bottlenecks,''
\newblock in {\em Conference on Computer Vision and Pattern Recognition
  (CVPR)}, 2018, pp. 4510--4520.

\bibitem{perazzi2016benchmark}
Federico P., Jordi PT, Brian M., Luc Van~Gool, Markus Gross, and Alexander
  Sorkine-Hornung,
\newblock ``A benchmark dataset and evaluation methodology for video object
  segmentation,''
\newblock in {\em Conference on Computer Vision and Pattern Recognition
  (CVPR)}, 2016, pp. 724--732.

\bibitem{pont20172017}
Jordi PT, Federico P., Sergi C., Pablo Arbel{\'a}ez, Alex SH, and Luc Van~Gool,
\newblock ``The 2017 davis challenge on video object segmentation,''
\newblock {\em arXiv:1704.00675}, 2017.

\bibitem{xu2018youtube}
Ning Xu, Linjie Yang, Yuchen Fan, Dingcheng Yue, Yuchen Liang, Jianchao Yang,
  and Thomas Huang,
\newblock ``Youtube-vos: A large-scale video object segmentation benchmark,''
\newblock {\em arXiv:1809.03327}, 2018.

\bibitem{cheng2021modular}
Ho~Kei Cheng, Yu-Wing Tai, and Chi-Keung Tang,
\newblock ``Modular interactive video object segmentation: Interaction-to-mask,
  propagation and difference-aware fusion,''
\newblock in {\em Conference on Computer Vision and Pattern Recognition
  (CVPR)}, 2021, pp. 5559--5568.

\bibitem{li2020fast}
Yu~Li, Z.~S., and Y.~S.,
\newblock ``Fast video object segmentation using the global context module,''
\newblock in {\em European conference on computer vision (ECCV)}, 2020, pp.
  735--750.

\bibitem{yang2020collaborative}
Zongxin Yang, Yunchao Wei, and Yi~Yang,
\newblock ``Collaborative video object segmentation by foreground-background
  integration,''
\newblock in {\em European conference on computer vision (ECCV)}, 2020, pp.
  332--348.

\end{thebibliography}
